\documentclass[letterpaper, 10 pt, conference]{ieeeconf}  

\IEEEoverridecommandlockouts                              

\usepackage{soul}
\usepackage{graphics} 
\usepackage{epsfig} 
\usepackage{mathptmx} 
\usepackage{times} 
\usepackage{amsmath,amssymb}
\usepackage{subfig}
\usepackage[T1]{fontenc} 
\usepackage{cite}
\usepackage{booktabs}
\usepackage{amsmath,amssymb,amsfonts}
\usepackage{algorithmic}
\usepackage{setspace}
\usepackage{graphicx}
\usepackage{textcomp}
\usepackage{xcolor}
\usepackage[T1]{fontenc}
\usepackage{aecompl}
\usepackage{multirow}
\usepackage{threeparttable}
\usepackage{tabularx}
\usepackage{mathtools}
\usepackage[super]{nth}
\usepackage[free-standing-units=true]{siunitx}
\usepackage[colorinlistoftodos,prependcaption,textsize=tiny]{todonotes}
\usepackage{balance}

\usepackage[ruled,vlined]{algorithm2e}
\newcommand{\Rom}[1]{\uppercase\expandafter{\romannumeral #1\relax}}
\newcommand{\rom}[1]{\expandafter{\romannumeral #1\relax}}

\usepackage{array}
\newcommand{\PreserveBackslash}[1]{\let\temp=\\#1\let\\=\temp}
\newcolumntype{C}[1]{>{\PreserveBackslash\centering}p{#1}}
\newcolumntype{R}[1]{>{\PreserveBackslash\raggedleft}p{#1}}
\newcolumntype{L}[1]{>{\PreserveBackslash\raggedright}p{#1}}

\newcommand{\fref}[1]{Fig. \ref{#1}}
\newcommand{\sref}[1]{Section \ref{#1}}
\newcommand{\tref}[1]{Table \ref{#1}}
\newcommand{\alref}[1]{Algorithm \ref{#1}}

\makeatletter
\let\NAT@parse\undefined
\makeatother

\usepackage[colorlinks]{hyperref}
  \hypersetup{
    citecolor=blue,
    linkcolor=blue,   
    urlcolor=blue}

\author{Xiao Lin$^{1}$ and Chen Wang$^{2}$
\thanks{$^{1}$Xiao Lin is with the College of Computing, Georgia Institute of Technology, Atlanta, GA 30332, USA. {\tt\small xiaol@sairlab.org}}%
\thanks{$^{2}$Chen Wang is with the Spatial AI \& Robotics Lab at The Department of Computer Science and Engineering, State University of New York at Buffalo, NY 14260, USA. {\tt\small chenw@sairlab.org}}%
}
\begin{document}

\title{\LARGE \bf
AirLine: Efficient Learnable Line Detection with Local Edge Voting
}

\thispagestyle{empty}
\pagestyle{empty}

\maketitle

\begin{abstract}
Line detection is widely used in many robotic tasks such as scene recognition, 3D reconstruction, and simultaneous localization and mapping (SLAM). Compared to points, lines can provide both low-level and high-level geometrical information for downstream tasks. In this paper, we propose a novel learnable edge-based line detection algorithm, AirLine, which can be applied to various tasks. In contrast to existing learnable endpoint-based methods, which are sensitive to the geometrical condition of environments, AirLine can extract line segments directly from edges, resulting in a better generalization ability for unseen environments. To balance efficiency and accuracy, we introduce a region-grow algorithm and a local edge voting scheme for line parameterization. To the best of our knowledge, AirLine is one of the first learnable edge-based line detection methods. Our extensive experiments have shown that it retains state-of-the-art-level precision, yet with a $3-80\times$ runtime acceleration compared to other learning-based methods, which is critical for low-power robots.
\end{abstract}

\section{Introduction}

With the growing need for autonomous robots in recent years, algorithms enabling robots to sense and interpret the real world through vision are becoming increasingly popular and demanded. Tasks like scene recognition \cite{Wang2020VisualLearning} and simultaneous
localization and mapping (SLAM) \cite{qiu2022airdos} are crucial for many applications such as autonomous driving \cite{Caesar2020Nuscenes:Driving} and reconstruction \cite{Kolmogorov2002Multi-cameraCuts}. To achieve matching and mapping in images from different views of the same scene, the prevailing solutions are often based on interest points \cite{Sarlin2020SuperGlue:Networksb,Detone2018SuperPoint:Description} with descriptors, or descriptive lines \cite{Vakhitov2019LearnableSLAM,Nguyen2021LS-Net:Detector}. A powerful point or line detector can provide a more consistent and informative result, significantly boosting the accuracy of image matching \cite{Ma2021ImageSurvey}, camera localization \cite{Brahmbhatt2018Geometry-AwareLocalization}, and environment mapping \cite{Amigoni2010AnRobots}.

Despite the promising progress, the performance of line detection is still unsatisfactory for real-time robotic tasks. On the one hand, traditional hand-crafted methods like LSD \cite{GromponeVonGioi2010LSD:Control} and Hough Transform \cite{Duda1972UsePictures}  with canny edge detection \cite{Canny1986ADetection} are efficient but sensitive to illumination and heavily rely on manual parameter tuning, often yielding unstable results and lacking generalization ability. They are computationally favorable but not adaptive or learnable, which might result in inconsistent detection in varying environments. Some recent traditional methods like \cite{Yu2020PLSD:Approach} outperform LSD but they still have a limited perceptual ability and adaptivity.

On the other hand, learning-based line detection methods are relatively robust but often computationally expensive \cite{wang2023pypose}.
For example, LCNN \cite{Zhou2019End-to-endParsing} detects line junctions and then learns to connect lines from points. Though the accuracy is appreciable, the inference speed is extremely slow. The state-of-the-art (SOTA) model LETR \cite{Xu2021LineEdges} achieves higher speed, but it is still not efficient enough to run in real-time.
Another learning-based approach is \cite{Zhao2022DeepDetection}, which performs infinitely-long semantic line detection with Hough Transform and deep-learned models. However, it is not suitable for line segment detection since Hough Transform can only detect infinitely long lines and may introduce potential errors by the limited discrete parametric space.

\begin{figure}[t]
\centering
\includegraphics[width=\linewidth]{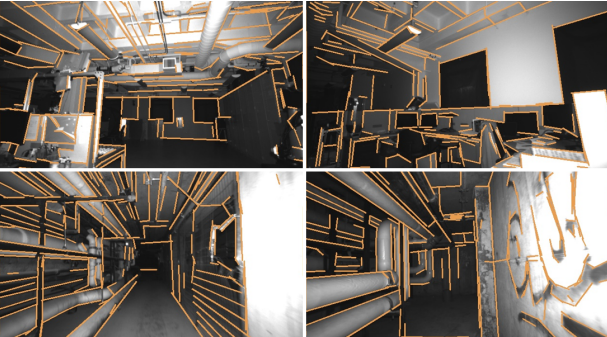}
\caption{AirLine is a lightweight line segment detector designed for robotic tasks. It succeeds in detecting most structural lines and is robust to viewport change and texture noise as the camera moves from scene to scene.}
\label{fig:sample}

\end{figure}
\begin{figure*}[t]
\centering
\includegraphics[width=\linewidth]{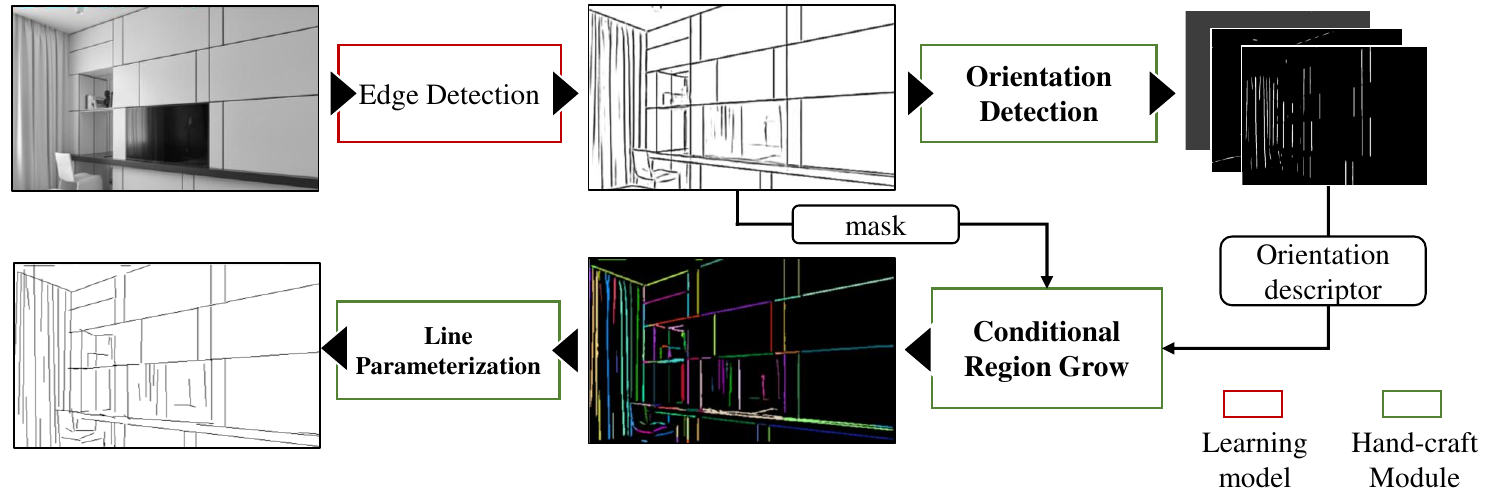}
\caption{The architecture of AirLine, which is extremely simple and only consists of four modules including edge detection, orientation detection, conditional region-grow, and line parameterization.}
\label{fig:structure}
\end{figure*}

One interesting phenomenon we observed is that most existing learnable line detection methods like LCNN are endpoint-based, which are sensitive to the geometrical conditions of the environments. However, not all lines contain significant endpoints, and there could be a large number of endpoints present in the environment, which can drastically increase computational expense.
To solve this problem, we introduce a learnable edge-based line detection method, AirLine, which can extract line segments directly from edges, resulting in better stability.
Specifically, we propose a novel edge voting scheme, based on a conditional region-growing strategy, combining the efficiency of classical methods and the perceptual ability of learning-based methods. This makes AirLine much faster than other learning-based methods and more robust and accurate than classical methods.

In summary, the main contributions of this paper are:
\begin{itemize}
\item We propose AirLine, a fast, robust, and accurate hybrid edge-based line detection pipeline without explicit end-points detection, where each module is plug-and-play and can be easily substituted by other better methods.
\item We design a new edge-to-line conversion scheme with local edge voting, which avoids the sensitivity and low efficiency of endpoint-to-line methods, as well as the line overlapping problem and low quality of the traditional edge-to-line methods based on Hough Transform.
\item We propose a new line segment detection metric based on pixel-level evaluation emphasizing edge-line consistency. We also demonstrate that AirLine retains state-of-the-art-level precision yet with a $3-80\times$ run-time acceleration. To the best of our knowledge, AirLine is the first learnable edge-descriptor-based line detection method. The source code is released at \url{https://github.com/sair-lab/AirLine}.
\end{itemize}

\section{Related Work}

\subsection{Hand-crafted Algorithms}

Line segment detector (LSD) \cite{GromponeVonGioi2010LSD:Control}, a classic hand-crafted line detector, remains the first preference for most line detection tasks in the robotics field. It relies on a region-growing strategy and runs extremely fast even without GPU to achieve real-time detection. In LSD, pixels with similar image gradients are grouped together to form a region and then converted to line segments by an approximated rectangle.
As a step forward, Cho \textit{et al.} \cite{Cho2018ADetection} not only utilized gradient but also other information such as brightness and gradient intensity to further improve the performance. Some other variants like PLSD \cite{Yu2020PLSD:Approach} developed line merging strategies but sometimes tend to over-segment.
An obvious limitation of the hand-crafted approaches is that the detection is based on low-level information like image gradient and brightness, leading to the unawareness of higher-level semantic information.

\subsection{Learning-based Algorithms}

A classic learning-based algorithm is LCNN \cite{Zhou2019End-to-endParsing}, which detects endpoints and learns to connect them and form line segments, making no use of edges and relying heavily on endpoint detection. The learned network pairs the detected endpoints to form line segments when the score exceeds a certain threshold.
LCNN has achieved high-quality line segment detection on widely-used datasets such as Wireframe \cite{Huang2018LearningEnvironments} and YorkUrban \cite{Denis2008EfficientImagery}. However, if a large number of endpoints exist, the inference time increases significantly, making stable real-time detection not possible. HAWP \cite{Xue2020Holistically-AttractedParsing} optimizes efficiency based on LCNN, achieving similar results in less time, but is still too computationally expensive to support real-time detection.
Unlike previous methods, LETR \cite{Xu2021LineEdges} detects lines as entities and represents them similar to the diagonal line of a bounding box produced by a vectorized line segment predictor, converted from the box predictor in DETR \cite{Carion2020End-to-EndTransformers}. This method has achieved state-of-the-art results on the Wireframe dataset under certain metrics, enabled end-to-end training of line segment detection, and reduced run-time variance compared to previous methods.

\subsection{Hybrid Algorithms}

Zhao \textit{et al.}\cite{Zhao2022DeepDetection} proposed a hybrid pipeline to achieve semantic line detection with Hough Transform. By voting edge-like information, a line is determined by the highest-voted angle and position. A differentiable Hough Transform is used to make the pipeline learnable. Although Hough Transform works well for a few straight lines, it offers a view too global for dense line segment detection. There are also learnable versions of LSD like LSDNet \cite{lsdnet} improving LSD by introducing learned line heat maps and orientation maps, but it is based on roughly predicted orientation instead of only edges. AirLine extracts orientation descriptor directly from edges without the need of extra predictors. 

Another edge-based hybrid approach is Pixel Orientation Estimation (POE) proposed by Liu \textit{et al.}~\cite{liu2022novel}, who designed a line detector based on edge grouping with estimated orientation and has shown its effectiveness on images with speckle noise.
POE operates on binary edge maps and use CPU for orientation estimation and segment formation. Our method, AirLine, further extends it by using a feature-based conditional region-grow and GPU-based convolution for voting. Additionally, AirLine accepts grouping conditions based on edge descriptors other than orientations and exploit GPU computing for efficient inference.

\begin{figure}[t]
\centering
\includegraphics[width=\linewidth]{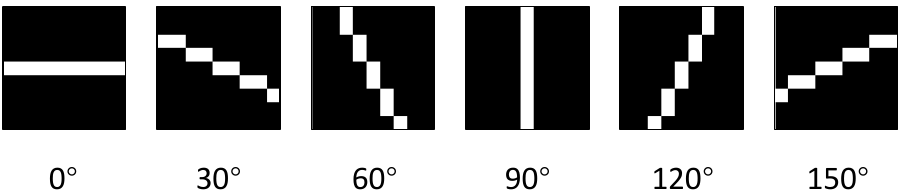}
\caption{The voting convolution kernels in the orientation detection module. Each channel contains a 1-pixel-wide line of a specific orientation.}
\label{fig: OD kernels}
\end{figure}

\begin{figure}[t]
\vspace{-10pt}
\centering
\subfloat[Detect Orientation]{\includegraphics[width=0.3\linewidth]{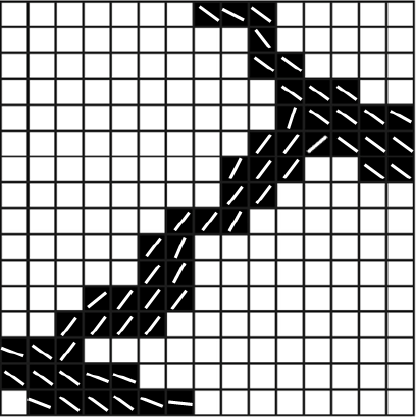}
\label{Fig: micro view 0}
}
\subfloat[CRG]{\includegraphics[width=0.3\linewidth]{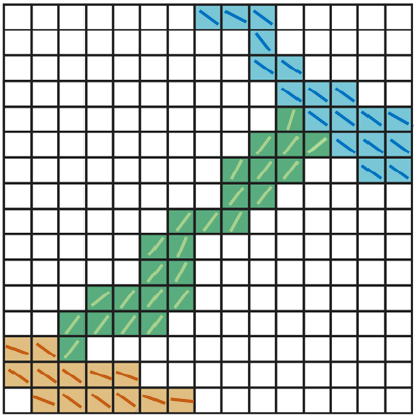}
\label{Fig: micro view 1}
}
\subfloat[Parameterization]{\includegraphics[width=0.3\linewidth]{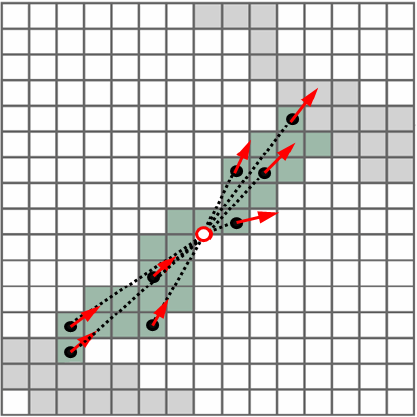}
\label{Fig: micro view 2}
}

\caption{A hand-craft illustration of how the orientation detection and conditional region-grow help convert edges to lines. Figure (a) is an orientation map detected on the thresholded edge; Figure (b) is the result of the conditional region-grow based on detected orientation; Figure (c) demonstrates the process of line parameterization: the white-red dot denotes the center $m'$ of the region and red arrows denote the tangent vectors $\mathbf{v}$ from each pixel in the region to $m'$.}
\end{figure}

\section{Methodology}

As shown in \fref{fig:structure}, AirLine is a learnable edge-based line detector with hybrid architecture that is composed of four modules including edge detection, orientation detection, conditional region-grow, and line parameterization. We next present their motivation and detailed process, respectively.

\subsection{Edge Detection}

The edge detection module is used to identify potential edges making up line segments. 
To this end, we employ an image-to-image network to predict edges as binary images. This module plays a significant role in determining the quality of lines, as it plays a role to filter out noisy textures and ensures the continuity of edges. In our experiments, we found that the detected edges do not need to be strictly straight as long as they are distinguishable and continuous, because our proposed conditional region-grow strategy presented in \sref{sec:CRG} is able to detect any straight, continuous edges. 

Specifically, we adopt the U-Net \cite{Ronneberger2015U-net:Segmentation} as our edge detector. 
To improve the continuity of the detected edges, we impose a weak penalty on the false positive predictions with a weighted masked binary cross entropy (BCE) loss $L_{\text{edge}}$:
\begin{equation}\label{eq:edge-loss}
    L_{\text{edge}} =  ((1-w)+w\cdot\tau_r(\mathbb{I}_{\mathbf{Y}=1})) \operatorname{BCE}(\mathbf{X},\mathbf{Y}),
\end{equation}
where $w$ is a weight for mask, $\textbf{X}$ is a prediction, $\textbf{Y}$ is ground truth, $\tau_r$ is an dilation operator with radius $r$, and $\mathbb{I}_{\mathbf{Y}=1}$ is a mask where its value is $1$ if the corresponding position of $\mathbf{Y}$ is also $1$, otherwise $0$.
The experiments show that this loss function stabilizes line detection and improves generalization for data outside the training distribution.

\subsection{Orientation Detection}\label{sec: Orientation Detector}

After performing edge detection, we design an orientation detection module to predict the slope of the edges.
The orientation detector is simply a convolution layer with $N$ hand-crafted kernels for each channel. Specifically, the $n^{\text{th}}$ channel of the kernel contains a 1-pixel-wide line at an angle of $n \times\frac{180^{\circ}}{N}$ ($n=0,\cdots,N-1$) drawn with value 1 and the rest pixels being 0, as shown in \fref{fig: OD kernels}. By applying the orientation detection to the edges and then performing an $L_2$ normalization \eqref{eq: orientation convolution}, we can easily generate a $N$-channel orientation descriptor $D_{\mathbf{x}}$, of which each channel describes the probability of the pixel belongs to a particular orientation:
\begin{equation}\label{eq: orientation convolution}
    D_{\mathbf{x}}=\phi(\mathrm{Conv}(\mathbf{X}, \mathbf{k})),
\end{equation}
where $D_\textbf{x}$ is the output descriptor map, $\phi$ is an $L_2$ normalization function, $\mathrm{Conv}$ is a convolution, and $\mathbf{k}$ is the handcraft $N$-channel kernel. The convolution is applied to the thresholded edge map, resulting in an orientation descriptor map as visualized in \fref{Fig: micro view 0}.

Initially, we attempted to directly predict the edge orientation map with learnable models such as U-Net, but it proved to be difficult to train and of poor precision. 
By employing this handcraft convolution, we can precisely compute orientation extremely fast in parallel, especially with libraries such as PyTorch \cite{Paszke2019PyTorch:Library}. The kernels' size and number can be adjusted according to the different applications. Larger kernel sizes and more kernels might result in excessive sensitivity. In our experiments, $N=6$ provides the best overall performance in terms of accuracy and efficiency.

\begin{algorithm}[t]
\caption{Conditional Region-grow (CRG)}\label{alg:CRG}
$G_\mathbf{X} \gets \phi $\\
\ForEach{\textnormal{unused edge pixel} p \textbf{in} $\mathbf{X}$}{
    $R \gets \phi$\\
    $F \gets \phi$\\
    $D_R^{\text{avg}} \gets D_p$\\
    \textbf{add} $\eta(p)$ \textbf{to} $F$\\
    \While{$F \neq \phi$} {
        \ForEach{\textnormal{unused pixel $p'$ \textbf{in} $F$}}{
            $S$ $\gets$ $D_{p'}$ $\cdot$ $D_R^{\text{avg}}$\\
            \If{$S \geq T$}{
                \textbf{mark} $p'$ \textbf{as} used\\
                \textbf{update} $D_R^{\text{avg}}$ \textbf{with} $D_{p'}$\\
                \textbf{add} $p'$ \textbf{to} $R$  \\
                \textbf{add} $\eta(p')$ \textbf{to} $F$ 
            }
        }
    }
    \If{$\mid R\mid  > m$}{
        \textbf{add} $R$ \textbf{to} $G_\mathbf{X}$
        }
}
\Return $G_\mathbf{X}$
\end{algorithm}

\subsection{Conditional Region-grow}\label{sec:CRG}

We next introduce a conditional region-grow (CRG) algorithm to group continuous straight edges and output edge groups which will be used for line parameterization.
In \alref{alg:CRG}, we first define $\eta(p)$ as an operator that returns 8 adjacent pixels around the pixel $p$. Then we declare $F$ as a list to store the frontier pixels during searching, and a threshold $m$ to skip regions with too few pixels.

\begin{figure}[t]
\centering
\includegraphics[width=\linewidth]{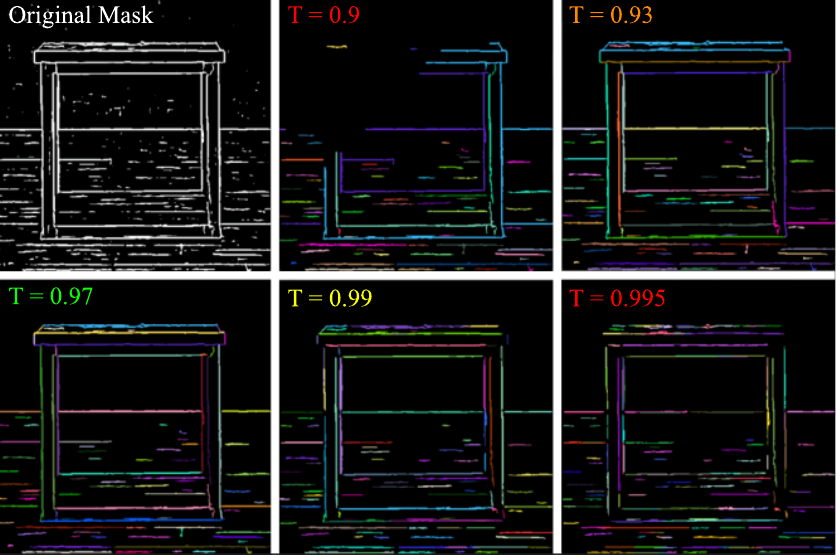}
\caption{This figure gives a straightforward showcase of how similarity threshold $T$ in \alref{alg:CRG} affects the result edge segmentation. A high $T$ means that the conditional region-grow is sensitive to descriptor change. In this example input, $T=0.97$ turns out to be the best threshold to keep the continuity of edges and tend not to be overgrown at corners.}
\label{Fig: threshold demo}
\end{figure}

Different from the region-grow algorithm in LSD, which grows the region by image gradients, we grow each region along edge pixels with similar orientation descriptors produced by orientation detection to identify groups of straight edges as visualized in \fref{Fig: micro view 1}.
As a result, our algorithm can stop growing edges at turning corners.
The criterion $C_p$ of whether a pixel $p$ will be merged into a region $R$ can be represented as the following expressions:
\begin{align}\label{eq:CRG similarity}
       C_p &=\begin{cases}
            \text{True}  & S_p \geq T,\\
            \text{False} & S_p < T,\\
            \end{cases}\\
    S_p & =D_{p} \cdot D_R^{\text{avg}},
\end{align}
where $S_p$ describes the similarity of descriptor for pixel $p$ and averaged grown region descriptor $D_R^{\text{avg}}$ and $T$ is a user-defined similarity threshold. If $C_p$ is true, the pixel $p$ will be grown as part of a region $R$ and then the averaged region descriptor $D_R^{\text{avg}}$ is updated with $D_{p}$. The threshold $T$ is subject to change for a suitable sensitivity to orientation changes as demonstrated in \fref{Fig: threshold demo}, which can be used for edge-to-curve detection as well.

\subsection{Line Parameterization}

The final step of AirLine is to parameterize the grown straight edge groups into line segments by local edge voting. Specifically, we aim to extract the centers, tangent vectors, and endpoints for all potential lines. Though it is possible to directly calculate the tangent vectors with the regional mean orientation descriptor $D_R^{\text{avg}}$, in the experiments we found that parameterizing with exact pixel coordinates is more precise. 

We next show how to parameterize a line with all grown pixels in a group $G$. 
In contrast to the traditional Hough transform, which uses all edge pixels across the image to vote for all lines simultaneously, we instead convert group-wise edges into lines, which can avoid interference between different groups. In detail, we first determine the center of mass $\mathbf{m}$ as an anchor point on the line; Then vote a tangent vector $\mathbf{\hat{v}}$ to express the orientation of the line; Finally, the endpoints are obtained from the farthest pixel to the center.
Therefore, for each line, the parameters can be calculated as: 
\begin{subequations}\label{eq:graph_edg_q}
    \begin{align}\label{eq:graph_edg_k}
        \mathbf{m} &= (\bar{x},\bar{y}), \quad (x, y) \in G\\
        \mathbf{\hat{v}} &=\phi\left(\sum_{(x_i, y_i) \in G}{{x}_i-\mathbf{m}_{x}}, \sum_{(x_i, y_i) \in G}{y_i-\mathbf{m}_{y}}\right),\\
        \mathbf{p} &=\pm (\mathbf{\hat{v}} \cdot \mathbf{v}_{\max})\mathbf{\hat{v}}+\mathbf{m},
    \end{align}
\end{subequations}
where $x$ and $y$ are coordinates of the grown edges, $\bar{\cdot}$ indicates the average operation, $\phi$ is an $L_2$ normalization function, $\mathbf{\hat{v}}$ is the voted tangent vector, $\mathbf{v}_{\max}$ is the longest vector from the center to pixels, and $\mathbf{p}$ is the endpoint of the line, which is the final output.
\fref{Fig: micro view 2} illustrates a visualization reference for this step. At this point, the line segments have been successfully extracted using the aforementioned modules.

\begin{figure}
\centering
\includegraphics[width=0.95\linewidth]{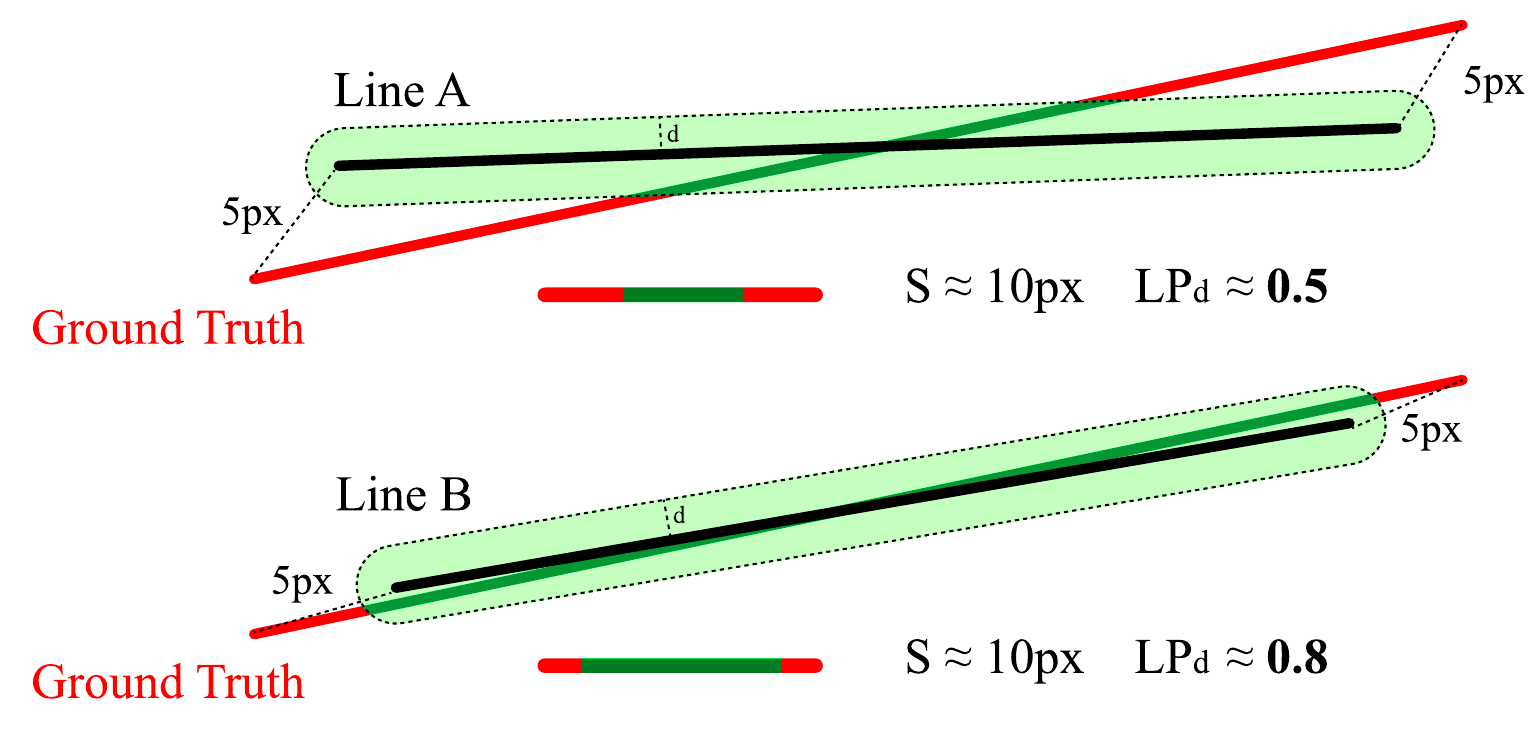}
\caption{Endpoint-based metric cannot well evaluate the two detected lines which have the same endpoint shifts S, while pixel-level precision is a better indicator for line detection.}
\label{fig: metric visualization}
\end{figure}

\subsection{Evaluation Metric}

Prior line segment detection methods, e.g., \cite{Xu2021LineEdges, Zhou2019End-to-endParsing} have typically been evaluated using the endpoint precision $_\text{s}$\text{AP}, which defines a criterion for true positive detection:
\begin{equation}\label{eq:sAP}
    \min_{(p_n, p_m) \in L} (\parallel{\Bar{p_1}-p_n}\parallel^2_2+\parallel{\Bar{p_2}-p_m}\parallel^2_2)\leq E,
\end{equation}
where $\Bar{p_1}$ and $\Bar{p_2}$ are the ground truth endpoints, $p_n$ and $p_m$ are detected endpoints of lines $L$, and $E$ is a user-defined threshold. However, it is not reasonable for many downstream applications due to three reasons:

\textit{1)} $_\text{s}$\text{AP} disregards measurements like line length and orientation, which is crucial for many downstream tasks such as pose estimation.
In many cases, long lines weigh much more than short ones in terms of semantic significance as well as geometrical information. However, as shown in \fref{fig: metric visualization}, a line with a relatively large angle error could have a similar endpoint error as a line with the correct orientation, therefore the endpoint precision cannot faithfully reflect the effectiveness of line detection.

\textit{2)} Some hand-crafted datasets like Wireframe\cite{Huang2018LearningEnvironments} and YorkUrban \cite{Denis2008EfficientImagery} label lines in an inconsistent way, i.e., some lines are labeled in one image while similar lines are ignored in others. This indicates that line detection is a subjective task and annotators have a different definition for line significance. Therefore, we argue that there should be a grace tolerance to ``false positive'' line detection to mitigate the effects of the subjective factors.

\textit{3)} A relatively low score on endpoint precision does not necessarily indicate poor performance. For example, LSD is one of the most popular line detectors, but it has an extremely low $_\text{s}$\text{AP} score because it tends to detect separated segments, though the detected shorter lines are accurate.

Thus, we concluded that pixel-level line coverage precision could better reflect the quality of line detection.
To solve the above problems and provide a useful perspective in the line detection task, we propose a new metric, e.g., Line Precision (LP$_\text{r}$) in \eqref{eq:crd}, to consider all reasonable lines:
\begin{equation}\label{eq:crd}
\text{LP}_\text{r}=\frac{\sum{\tau_\text{r}(\textbf{X})\otimes \textbf{Y}}}{\sum{\textbf{Y}}}.
\end{equation}
In this definition, $\tau_\text{r}$ denotes a dilation function with a tolerance radius $r$, and $\otimes$ is an element-wise multiplication operation. LP$_\text{r}$ is designed for calculating the percentage of ground truth pixels covered by the prediction within an error range. We include \fref{fig: metric visualization} for a simple and straightforward interpretation. Again, we designed the metric to evaluate true positives to fairly compare both learned methods and handcrafted methods that behave differently, as a lot of ``false positives'' are reasonable but not labeled in the dataset.
Take line B in \fref{fig: metric visualization} as an example, LP$_\text{d}$ is given by dividing the area of ground truth by its intersection with the expanded output area.
The two examples have similar $_\text{s}$\text{AP} scores given the same amount of endpoint offsets, however, line B is a result better than line A in terms of orientation and position.

\begin{table}[!t]
    \caption{Overall performance comparison.}
    \label{tab:Wireframe precisions}
    \centering
    \resizebox{\linewidth}{!}{
    \begin{tabular}{cccccccc}
      \toprule
      Method & LP$_0$ & LP$_1$ & LP$_2$ & LP$_3$ & LP$_5$ & LP$_{10}$ & FPS\\
      \midrule
        AirLine & \underline{18.22} & \underline{49.99} & \underline{65.63} & \underline{75.5} & \textbf{84.21} & \textbf{94.52} & \underline{24.3}\\
        LCNN & \textbf{21.35} & \textbf{53.09} & \textbf{66.54} & \textbf{75.73} & \underline{83.98} & \underline{94.03} & 0.9\\
        LETR & 15.07 & 41.37 & 57.85 & 69.32 & 78.86 & 90.96 & 4.5\\
        LSD & 12.79 & 39.13 & 49.4 & 55.38 & 61.4 & 70.79 & \textbf{34.5}\\
      \bottomrule
      
    \end{tabular}
    }
    
\end{table}

\begin{figure}[t]
\vspace{-10pt}
\centering

\subfloat[Performance Contour\label{fig: WireFrame a}]{\includegraphics[width=0.465\linewidth]{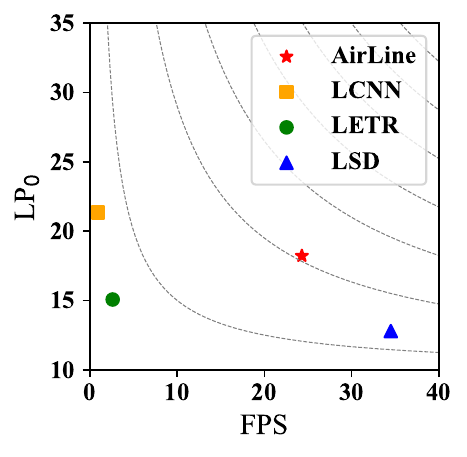}

\label{fig:Wireframe p1p10 precision}}
\subfloat[Line Precision\label{fig: WireFrame b}]{\includegraphics[width=0.48\linewidth]{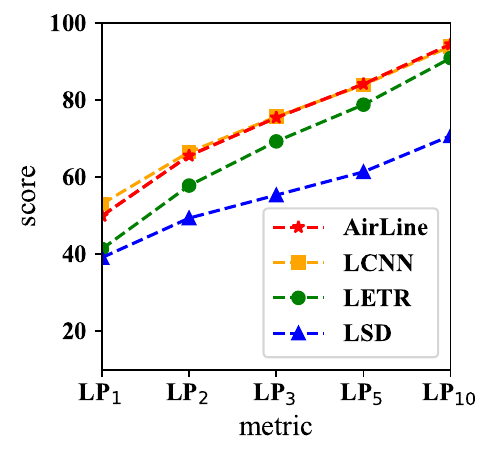}
}
\caption{Overall performance of AirLine. (a) is the performance contour (LP$_0$-FPS); The closer to the upper-right corner, the better the precision-speed performance. (b) presents line precision with different settings. LP$_0$ denotes different methods' percentage probability to strictly cover GT pixels without dilation. Curves in (a) are defined by $y=\frac{\text{LP}_0\cdot\text{FPS}}{x}+10$, we offset y-coordinate by 10 because it can be taken as a baseline that can be produced by some random line detector. }
\label{Fig: WF}
\end{figure}

\begin{figure*}[t]
\centering\includegraphics[width=1\textwidth]{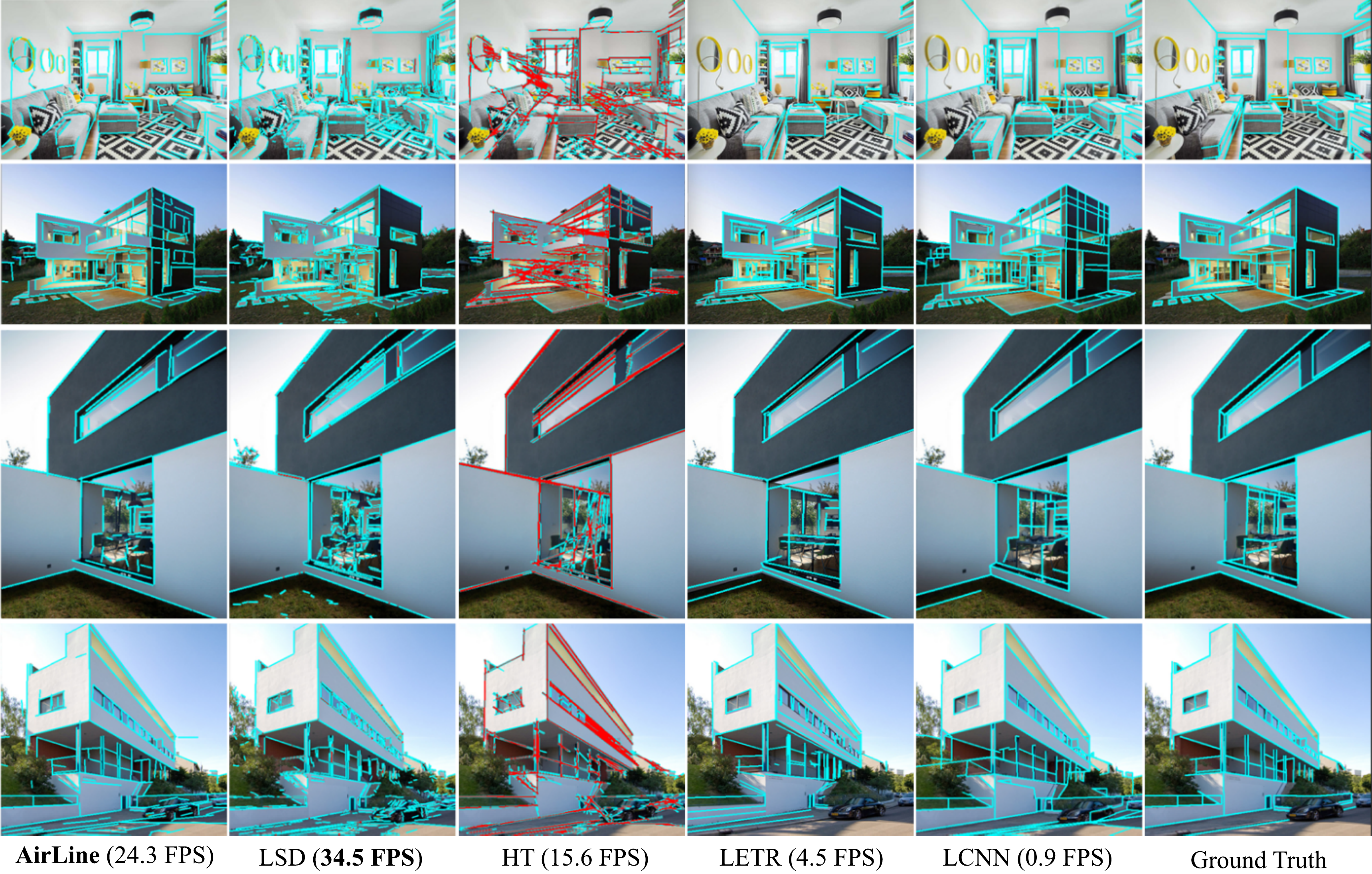}
\caption{The qualitative results on the Wireframe dataset \cite{Huang2018LearningEnvironments}. Images to examine line quality, noisiness, and generalization ability of different methods on images from the test set. *HT contains overlapped lines and we use red color to highlight duplicate detection, while little line overlaps exist in other methods. FPS is tested under the same resolution and hardware.}
\label{figure: Wireframe qualitative Comp}
\end{figure*}

\begin{figure}[t]
\centering

\subfloat[Performance Contour\label{fig: YorkUrban a}]{\includegraphics[width=0.465\linewidth]{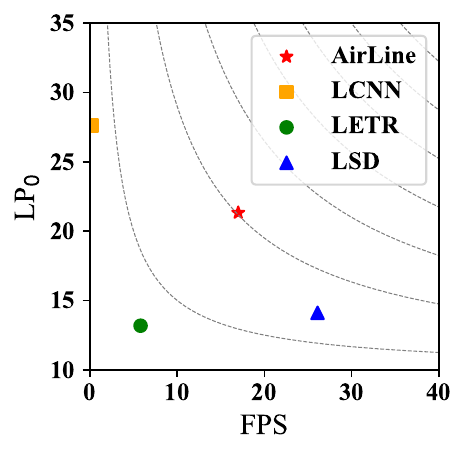}
}
\subfloat[Line Precision]{\includegraphics[width=0.48\linewidth]{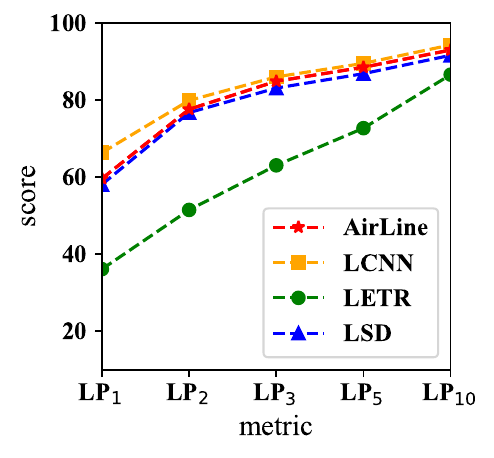}
}

\caption{Generalization Test. (a) and (b) are the performance counter and LP scores, respectively. In (b) the steepness of curves reflect the output stability: a less steep curve implies a more consistent score even given greater error. AirLine and LSD achieve the 1$^{\text{st}}$ and 2$^{\text{nd}}$ best generalization performance.
}
\label{Fig: YU}
\end{figure}

\section{Experiments}

\subsection{Implementation Details}

To train the U-Net edge detector, we used the masked BCE loss \eqref{eq:edge-loss} with weight $w=0.8$ and implemented the dilation function $\tau_r$ in the loss function using a transposed circle convolution with radius $r=5$. An Adam optimizer \cite{Kingma2015Adam:Optimization} with an initial learning rate $5\times 10^{-4}$ (reduced by multiplying 0.3 when reaching a milestone) is used for optimization. 

To make the edge detector equally sensitive to all orientations, we implemented data augmentation techniques including random rotation and brightness changes.

To create our hand-crafted orientation detector, we employed an unbiased PyTorch \cite{Paszke2019PyTorch:Library} Conv2d layer with kernels generated by OpenCV \cite{Bradski2000TheLibrary}. We ran the edge and orientation detector on the GPU and CRG and line parameterization on the CPU. The final output is a list of endpoints. We set the similarity threshold $T$ to 0.98 and a minimum of pixels $m$ for line parameterization to 15 in CRG (\ref{alg:CRG}).

\begin{figure*}[t]
\centering\includegraphics[width=0.99\textwidth]{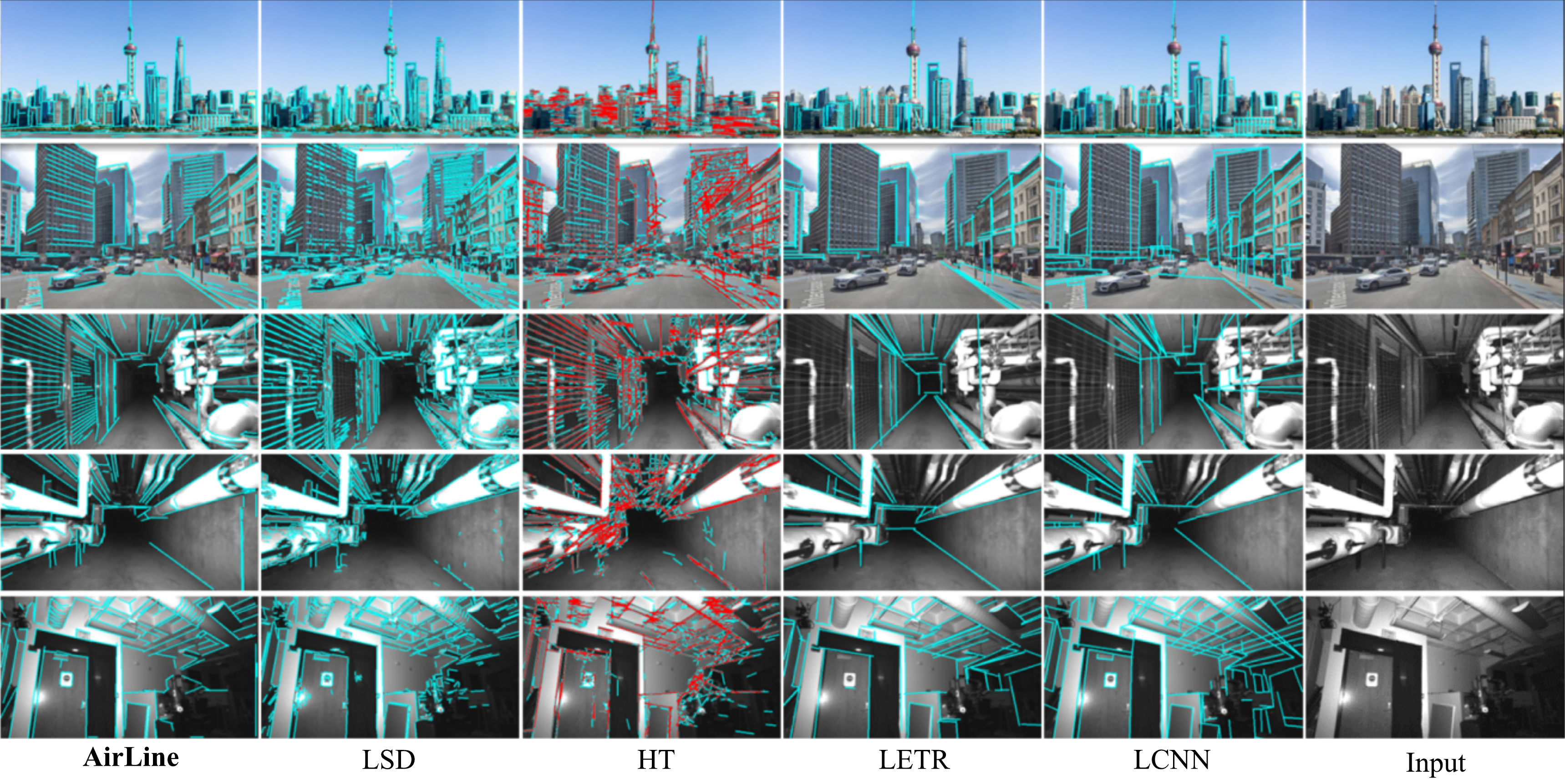}
\caption{Generalization test on online images and subterranean environments. We highlighted overlap detection in red as \fref{figure: Wireframe qualitative Comp}. 
The 1st, 2nd, and 3rd-5th rows are from city view, street view, and OIVIO dataset \cite{Kasper2019AIllumination}, respectively.}
\label{Fig. qualitative comparison2}
\end{figure*}

\subsection{Overall Performance}\label{sec:overall}

We first conduct a quantitative evaluation and compare with the state-of-the-art methods including LCNN \cite{Zhou2019End-to-endParsing}, LETR \cite{Xu2021LineEdges}, and LSD \cite{GromponeVonGioi2010LSD:Control} using the widely-used Wireframe \cite{Huang2018LearningEnvironments} dataset. It consists of a set of 5,000 images of housing structures, which is suitable for both training and accuracy tests. We present the LP scores and frames per second (FPS) on the test set in \tref{tab:Wireframe precisions}, in which AirLine achieves comparable accuracy with the best-scored method LCNN but is of much higher runtime efficiency, i.e., roughly 25$\times$ faster than LCNN.
For better visualization, we show the performance contour with LP$_0$ (the most strict metric) in \fref{fig: WireFrame a} and the full LP scores in \fref{fig: WireFrame b}.
As can be seen, AirLine achieves the best overall performance in terms of balanced accuracy and efficiency with a large margin than other methods with appreciable quality (as shown in \fref{figure: Wireframe qualitative Comp}).
It is noticeable that LSD achieves the 2$^{\text{nd}}$ best overall performance in \fref{fig: WireFrame a}, which also supports the fact that LSD is still preferred nowadays over other more recent methods, e.g. LCNN, in robotic tasks.

\subsection{Quantitative Generalization Test}

We further test the generalization performance of AirLine by testing the model trained on Wireframe dataset on the YorkUrban \cite{Denis2008EfficientImagery} dataset.
Different from Wireframe which primarily contains images for demonstration, YorkUrban offers a larger variety of live shots across different environments, making it ideal for quantitative generalization tests.
As shown in \tref{tab: YorkUrban precisions} and \fref{Fig: YU}, although AirLine is the 2$^{\text{nd}}$ best in terms of accuracy, it still possesses the most notable balance between accuracy and speed, i.e., 80$\times$ faster than LCNN.
This exceptional speed and efficiency make AirLine a top-performing method for learnable real-time line detection, which is crucial for on-field robotics.
Note that LSD achieves the 2$^{\text{nd}}$ best overall performance and ranks above LCNN and LETR in \fref{fig: YorkUrban a}, which further explains the popularity of LSD in robotic tasks and the rationality of the overall performance contour compared to sole accuracy.

\begin{table}[t]
    \caption{Generalization test.}
    \label{tab: YorkUrban precisions}
    \centering
    \resizebox{\linewidth}{!}{
    \begin{tabular}{cccccccc}

    \toprule
    Method & LP$_0$ & LP$_1$ & LP$_2$ & LP$_3$ & LP$_5$ & LP$_{10}$& FPS \\
    \midrule
    AirLine & \underline{21.33} & \underline{59.66} & \underline{77.66} & \underline{84.95} & \underline{88.58} & \underline{93.06}& \underline{17.0}\\
    LCNN & \textbf{27.63} & \textbf{66.39} & \textbf{79.96} & \textbf{86.08} & \textbf{89.59} & \textbf{94.37} & 0.21\\
    LETR & 13.18 & 36.14 & 51.52 & 63.10 & 72.76 & 86.66 & 5.8\\
    LSD & 14.11 & 58.12 & 76.80 & 83.17 & 86.91 & 91.68 & \textbf{26.1}\\
    \bottomrule
    \end{tabular}
    }
\end{table}

\subsection{Qualitative Generalization Test}

Next, we present the qualitative generalization test using online images and the OIVIO dataset \cite{Kasper2019AIllumination}. Although OIVIO does not offer ground truth lines for evaluation, it is ideal for generalization tests, especially for robotic applications, as it provides low-textured subterranean environments and scenes with dynamic illumination, and it was used in the well-known DARPA SubT challenge \cite{Scherer:2022,ebadi2022present}. As shown in \fref{Fig. qualitative comparison2}, we found that the AirLine and LSD methods performed better than LETR and LCNN in generating detailed line detections. Furthermore, AirLine is preferred over LSD because it can detect a thick line as one, while LSD tends to detect two sides of it. Although LCNN produces lines of high precision, it is not suitable for tasks that require detailed line detection, as it tends to ignore short lines and connect two unrelated endpoints in complex environments. LETR, which performed the best under the metric mentioned in its original paper, was found to have the worst line quality in terms of precision, stability, and reasonable geometry. Additionally, we tested the Hough Transform \cite{Duda1972UsePictures} (from OpenCV) in conjunction with our edge detector as a baseline for edge-to-line methods. It produces a large number of overlapping lines, while AirLine and other methods have almost no overlapped detection.

\begin{table}[!t]
    \caption{Ablation study of orientation descriptor.}
    \label{tab:ABTB}
    \centering
    \resizebox{\linewidth}{!}{
    \begin{tabular}{ccccccc}

    \toprule
    CRG input & LP$_0$ & LP$_1$ & LP$_2$ & LP$_3$ & LP$_5$ & LP$_{10}$ \\
    \midrule
    6-channel OD & \textbf{21.33} & \textbf{59.66} & \textbf{77.66} & \textbf{84.95} & \textbf{88.58} & \textbf{93.06}\\
    2-channel OD & 16.95 & 46.59 & 62.14 & 71.91 & 80.28 & 90.72\\
    Image gradient & 13.70 & 37.43 & 50.11 & 58.86 & 67.48 & 80.15\\
    \bottomrule
    \end{tabular}      
    }
    \vspace{-10pt}
\end{table}

\subsection{Ablation Study}
In addition, we conducted an ablation study on the orientation descriptor, a critical component of AirLine that connects learnable modules with handcraft algorithms. As presented in \tref{tab:ABTB}, we tested and compared three different orientation detection (OD) as input for conditional region-grow: 6-channel orientation descriptor as proposed in \sref{sec: Orientation Detector}; 2-channel orientation descriptor, using only 0$^\circ$ and 90$^\circ$ kernels shown in \fref{fig: OD kernels}; and 2-channel image gradient. By comparing the LP scores produced by 6-channel OD and 2-channel OD, we can see that the 6 channels perform better, though 2 channels are theoretically sufficient to describe the horizontal and vertical edges. By comparing the 2-channel orientation descriptor to 2-channel image gradient, we can see that our orientation descriptor can support more precise pixel-level detection (higher LP$_0$ score) and better semantic awareness (higher LP$_{10}$ score, which means that detector can cover more lines given greater tolerance). Therefore, our orientation detector is crucial for accurate edge segmentation, and the ablation study confirms its effectiveness.

\begin{figure}[t]
\centering\includegraphics[width=\linewidth]{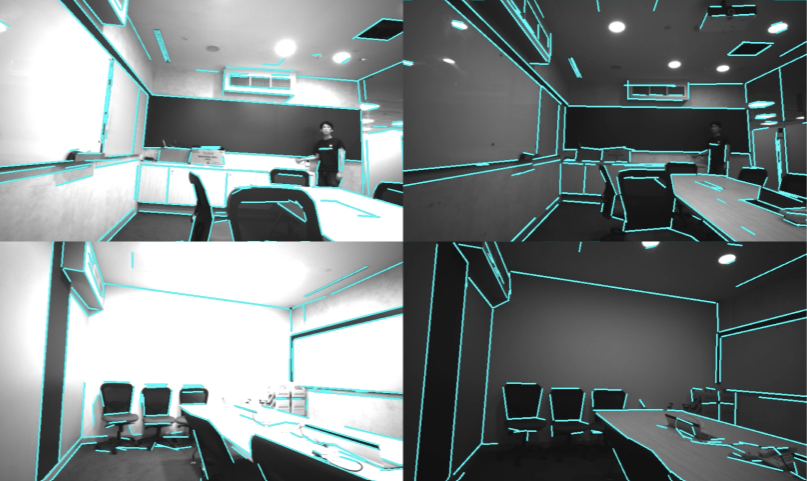}
\caption{Airline achieved robust detection with significant illumination changes in a live demo for office environments.}
\label{Fig: live demo}
\vspace{-10pt}
\end{figure}

\subsection{Live Demo}

We further present a live demo and show four snapshots in \fref{Fig: live demo} to demonstrate the robustness of AirLine.
The goal is to detect lines as consistently as possible given changing illumination and viewport conditions. We observed a stable line detection even though the brightness in the environment is changing frequently: The left two images present two over-exposure frames while the right presents much darker frames of the same scene. Existing methods could be largely interfered with by the illumination change and produce inconsistent line detection at the same place, while AirLine kept track of the most significant lines from the over-exposed frame to the other frames.
For better visualization, we suggest the readers watch the video attached to this paper.

\section{Conclusion}
We introduce AirLine, a novel line segment detection method featuring a hybrid architecture, demonstrating its capacity for effective generalization across various scenes, with a significant runtime boost compared to state-of-the-art methods. Specifically, we design a novel edge-to-line detection scheme, addressing the instability inherent in endpoint-to-line methods and the low efficiency of other learnable approaches. We further propose a new line segment detection metric prioritizing pixel-level accuracy, which is more relevant in many robotic tasks such as SLAM. AirLine achieves state-of-the-art precision coupled with a notable $3-80\times$ acceleration in runtime. Our comparisons and a live demonstration affirm AirLine's effectiveness in real-world scenarios and its robustness to environmental changes.

\balance

\bibliographystyle{./bib/IEEEtran}
\bibliography{root}

\end{document}